\documentclass[11pt]{article}
\usepackage{geometry}
\usepackage{coling2020}
\usepackage{times}
\usepackage{url}
\usepackage{latexsym}
\usepackage{microtype}
\usepackage{graphicx}
\usepackage{epstopdf} 

\usepackage{array}

\usepackage{multicol}

\usepackage{color,soul}

\usepackage{color}
\usepackage{amsmath}
\newcounter{todocnt}
\colingfinalcopy 

\title{DUTH at SemEval-2020 Task 11: BERT~with~Entity~Mapping~for~Propaganda~Classification \vspace{0.3cm}}

\author{
Anastasios Bairaktaris
\And 
Symeon Symeonidis \\
\\
Database and Information Retrieval research unit,\\
Department of Electrical and Computer Engineering,\\
Democritus University of Thrace, Xanthi 67100, Greece.\\
{\tt \{anasbair1,ssymeoni,avi\}@ee.duth.gr} \\
\And 
Avi Arampatzis  \\
}

\date{}

\begin{document}
\maketitle
\begin{abstract}
 This report describes the methods employed by the Democritus University of Thrace (DUTH) team for participating in SemEval-2020 Task 11: Detection of Propaganda Techniques in News Articles. Our team dealt with Subtask 2: Technique Classification. 
We used shallow Natural Language Processing (NLP) preprocessing techniques to reduce the noise in the dataset, feature selection methods, and common supervised machine learning algorithms. 
 Our final model is based on using the BERT system with entity mapping. 
 To improve our model's accuracy, we mapped certain words into five distinct categories by employing word-classes and entity recognition. 
\end{abstract}

\section{Introduction}\label{sec:intro}
\blfootnote{
    %
    %
    
    %
    %
This work is licensed under a Creative Commons 
Attribution 4.0 International Licence.\newline%
\hspace*{0.26cm}
Licence details: \url{http://creativecommons.org/licenses/by/4.0/}
    %
    %
    \vspace{0.2cm}
}
According to the Institute for Propaganda Analysis\footnote{\url{https://propagandacritic.com/}}, propaganda is an expression of an opinion or an action by individuals or groups deliberately designed to influence the opinions or the actions of other individuals or groups concerning predetermined ends.
With the rapid change that the world wide web has made, it is evident that the means available for propaganda to be spread are more than ever before. The fact that, nowadays, news outlets can reach out to millions of people through their websites or social media demonstrates how easy it is to manipulate people with propaganda techniques or fake news.
For example, political forecasts severely underperformed in predicting the results of the 2016 US presidential election and the United Kingdom European Union membership referendum (Brexit) as opposed to the consensus in social media, which is indicative of the new challenges that are upon us \cite{DBLP:journals/computer/HallTJ18}.

The SemEval-2020 Task 11: Detection of Propaganda Techniques in News Articles aims to produce models that can identify text fragments with various propaganda techniques. The first subtask is a binary sequence tagging task in which a model has to return the spans that contain at least one propaganda technique. The second subtask is a multi-class classification task in which given a text fragment and the article it occurs in, participants must classify the fragment into one of 14 different propaganda classes. More details on the Task can be found on the Task Description paper \cite{DaSanMartinoSemeval20task11}.

The rest of this paper is structured as follows.
Section~\ref{sec:related} outlines some previous studies of propaganda identification.
Section~\ref{sec:system} describes our approach, while Sections~\ref{sec:experiments} and \ref{sec:results} present experiments and results respectively. 
Conclusions are summarized in Section~\ref{sec:conclusions}.

\section{Background}\label{sec:related}

Barr\'{o}n-Cede\~no et al.~\shortcite{DBLP:conf/aaai/Barron-CedenoMJ19} presented \texttt{Proppy}, a publicly available real-time propaganda detection system that is used for online news. The system used four modules that include article retrieval, event identification, deduplication, and propaganda index computation. To organize the news based on their propagandistic content, they showed that when identifying propaganda, approaches that use word n-grams are less effective than those that use character n-grams and other style features. Additionally, Da San Martino et al.~\shortcite{EMNLP19DaSanMartino} introduced a new approach of analyzing propaganda that focuses on identifying fragments that contain propaganda techniques as well as their type, as opposed to addressing propaganda detection at the document level.

Rashkin et al.~\shortcite{DBLP:conf/emnlp/RashkinCJVC17} described the need for examining lexical features when trying to understand the differences between more and less reliable digital news sources. They studied the usefulness of linguistic morphology in different types of fake news such as propaganda, satire, and hoaxes. They also created a corpus of categorized news articles with labels such as propaganda, trusted, hoax, or satire. In another major study, 
Rashkin et al.~\shortcite{DBLP:journals/corr/abs-1910-09702} noted the importance of discovering relationships between different propaganda techniques. They hypothesized that finding common traits could prove helpful in classification tasks. In our approach, we investigated the effects of entity mapping in certain classes, and our conclusions are in line with Rashkin et al.~\shortcite{DBLP:conf/emnlp/RashkinCJVC17} concerning the existence of conceptual and linguistic relationships between propaganda techniques.

In recent years, there have been some significant landmarks in the NLP field. 
ELMo \cite{DBLP:conf/naacl/PetersNIGCLZ18}, ULMFiT \cite{DBLP:conf/acl/RuderH18}, OpenAI GPT \cite{Radford2018ImprovingLU}, and BERT \cite{DBLP:journals/corr/abs-1810-04805} are some large scale models that have massively improved the results in many NLP tasks. These systems provide models that have been pre-trained in massive corpora of unlabeled data and require fine-tuning in task-specific data. Although these systems offer excellent results, there is a need for further experimentation, as noted by Hua~\shortcite{DBLP:journals/corr/abs-1911-04525} which highlights BERT's shortcomings in real-world scenarios.

\section{Approach}\label{sec:system}


This section describes our approach to mapping certain words into five distinct categories by employing word-classes and entity recognition. It also introduces the BERT model which was employed for our final submission.

\subsection{Mapping the Dataset} \label{sssec:Mapping}

The main idea of our method was to investigate the relationship between different entities and whether they have relevant usage. This is demonstrated with the examples in Table~\ref{table:l-examp}. 
The Flag Waving technique is an example of how words that bear no similarity 
in a bag-of-words representation,
have the exact same semantic value for propaganda technique classification. In this example, `Soviet Union' and `Iran' have the same value (being both countries) for propaganda classification.

\begin{table}[!htbp]
		\begin{tabular}{ |c|c| } 
			\hline
			\bf Propaganda Technique & \bf Propaganda Extract \\
			\hline
			Flag Waving & \textit{`This is not the \textbf{Soviet Union}, this is not \textbf{Iran} or Riyadh this is \textbf{America}.'} \\ 
	
			Name\_Calling,Labeling &\textit{`\textbf{fascist} propaganda tropes.'}\\
	
			Slogans 
			&\textit{`Make America Great Again'}\\
			\hline
		\end{tabular}
	\caption{\label{table:l-examp}Samples from different labels}
\end{table}

The same applies for entities such as `communists' and `fascist' (political ideologies) and `Christians' and `Muslims' (religious groups). The hypothesis is that, for propaganda classification, when someone wants to attack another nation or a certain group through propaganda, it is less important which group or nation initiates or receives the attack. Thus, we made three lists that aim to reduce the noise in data that is produced from various countries, religious or political groups. We also made a list that contained different slogans to help with the Slogans category.

The lists we created are the following and can be found on github\footnote{\url{https://github.com/anasbair/SemEval2020-groups}}:
\begin{multicols}{2}
\begin{itemize}
	\item \textbf{List\_Countries:} The names of 255 countries as well as some variations such as `America' or `UK'.
	\item \textbf{List\_Religion:} 35 words that relate to religion such as `Catholic' and `Muslim'.
	\item \textbf{List\_Politics:} 23 words that relate to politics such as `Democrat' or `Republicans'.
	\item \textbf{List\_Slogans:} 41 slogans such as `War on Terror' or `Build the wall'.
\end{itemize}
\end{multicols}
We scanned the dataset for those instances and replaced them with the following tags:
\textbf{NATION}, \textbf{RELIGION}, \textbf{POLITICS}, and \textbf{SLOGANS}. 
The final results showed that this approach improved significantly the basic BERT model. 

\subsection{Named Entity Recognition}\label{sssec:EntityRec}

Named Entity Recognition is the process of identifying proper names and classifying them into categories such as persons, organizations, locations, etc. This process is vital for many NLP applications \cite{DBLP:conf/acl/PetasisVWPKS01}. Carrying on with our previous hypothesis, we also experimented with entity recognition. We noticed that in many instances of propaganda, there was a use of names of politicians that could be grouped to help the accuracy of our model. Although we experimented with many different entity groups/types such as Nationalities and Organisations, the best results came with the People entities.

To achieve this, we used SpaCy's\footnote{\url{https://spacy.io/}} named entity recognizer which has been trained on the OntoNotes 5 corpus \cite{DBLP:journals/ijsc/PradhanHMPRW07}. 
After the recognition, we replaced the entity with the PERSON tag. This approach yielded our best results in the Flag Waving category. 

\subsection{BERT - Bidirectional Encoder Representations from Transformers}
BERT is a language representation model that was introduced by Devlin et al.~\shortcite{DBLP:journals/corr/abs-1810-04805}. It stands for Bidirectional Encoder Representations from Transformers. BERT pre-trains deep bidirectional representations from text that has not been labeled. 
The fact that BERT is deeply bidirectional allows it to learn information during training from both sides of a token's context. The following two steps are involved in BERT.


The BERT model has been pre-trained in the BooksCorpus (800m words) \cite{DBLP:conf/iccv/ZhuKZSUTF15} and English Wikipedia (2,500m words). In the first step,  we fine-tuned the BERT model on different versions of the dataset that was provided by the organizers. 
BERT requires input data to be in a specific format. To mark the beginning, the [CLS] special token is used and for the separation or end of the sentences the [SEP] is used. The input is represented as:
%
$[\mathrm{CIS}]+\mathrm{text}+[\mathrm{SEP}]$
. 

The next step was to tokenize the propaganda extracts into tokens that match BERT's vocabulary. For tokenization, we used BERT's \texttt{BertTokenizer}. \texttt{BertForSequenceClassification}\footnote{\url{https://huggingface.co/transformers/model_doc/bert.html#bertforsequenceclassification}} which is the model that we used for fine-tuning. 
This BERT transformer has a sequence classification/regression head on top (a linear layer on top of the pooled output). According to the recommendations of Devlin et al.~\shortcite{DBLP:journals/corr/abs-1810-04805}, for training we used a batch size of 32, a learning rate of 2e-5, and the number of epochs was set to 4.

\section{Experimental Setup}\label{sec:experiments}

In this section, we describe the experimental setup of this study, providing information for the dataset and the parameters of machine learning algorithms, respectively.

\subsection{Dataset}

The organizers provided three datasets Training, Development, and Test. The training dataset consisted of 357 articles in text format, retrieved with Python's newspaper3k\footnote{\url{https://github.com/codelucas/newspaper/}}. 
For the second subtask, the organizers provided a text file with 6,129 propaganda text fragments, belonging to 13 categories, alongside their respective article id and the spans in which the technique was located in the article. 
The 13 categories/labels are shown in Table~\ref{table:bertfinal}.
The dataset is imbalanced since the Name\_Calling,Labeling and Loaded\_Language labels jointly constitute 50\% of the dataset.

\subsection{Pre-Processing} \label{sssec:prepro}
We tested various pre-processing techniques and by using the conclusions of Symeonidis et al.~\shortcite{DBLP:journals/eswa/SymeonidisEA18} we applied the following:
Remove Numbers, Remove Punctuation, Remove Symbols, Lowercase, and Replace all URL addresses normalizing them to `URL'.

We prefer not to remove stopwords due to the results of our previous work on SemEval-2019 Task 8: Fact Checking in Community Question Answering Forums 
\cite{DBLP:conf/semeval/BairaktarisSA19}.
In that work, we concluded that stopwords can prove important for certain tasks. For example, a common word such as `believe' can strongly indicate opinion and as such is useful.

\subsection{Machine Learning Model}\label{sssec:modelA}

Before using the BERT model for our final submission, we used standard machine learning methods for our experiments. We will briefly present these methods, which, just in two classes (Oversimplification and Flag Waving), performed better than the BERT model techniques on the development set. 

For the training of our classifiers, we used Python's Scikit-Learn library \cite{DBLP:journals/jmlr/PedregosaVGMTGBPWDVPCBPD11}. We split the given pre-labelled data into 2/3 training and 1/3 development set (2:1 ratio). After the split, the training set was shuffled, and tested a sequence of tuning parameters on the development set. When the test set was provided by Task organizers, we re-trained the classifiers into the total training set and tested on the organizers' test set.

\textbf{Vectorizer}: We compared three common vectorizers such as CountVectorizer, HashingVectorizer, and TfidfVectorizer. Finally, our selection was the TfidfVectorizer since it yielded the best results. 

\textbf{Classifiers}: We tested various classifiers and decided to use the following three: SGDClassifier, RidgeClassifier, and LinearSVC, as they yielded the best micro-averaged $F_{1}$ Results.
%

\section{Results}\label{sec:results}

This section summarizes our experimental results. 
Before our officially submitted run, we present some additional experiments.

\subsection{Machine Learning Model results}
In Table~\ref{table:modelA}, we present the results of our machine learning Baseline Model. The Baseline Model is with the RidgeClassifier, as described in Section~\ref{sssec:modelA}, since it yielded the best results on the training process.  We show the $F_{1}$-score of the classifier when the Baseline Model was trained with the mapped datasets that we described in Sections~\ref{sssec:Mapping} and \ref{sssec:EntityRec}. For the entity recognition we used a variety of entities such as persons, nationalities, organisations, countries, cities and locations. As we mentioned in Section \ref{sssec:EntityRec} the PERSON entity achieved the best results.
    
The Baseline Model achieved some notable results on the development set for two labels. 
In the Oversimplification label, the baseline model yielded a micro-averaged $F_{1}$ of 29\%, as opposed to the basic fine-tuned BERT which failed to recognize this class. Furthermore, for the Flag Waving label, the Baseline Model scored 1\% more than our best BERT model. However, as we can see in Table~\ref{table:testbert}, the BERT model performed better overall results and was selected for our final submission.
    
\begin{table}[ht]
	\begin{center}
		\begin{tabular}{ |c|c| } 
			\hline
			Technique &  $F_{1}$ Score \\
			\hline
			Baseline Model (Overall) & 46.37 \\ 
			\hline
			NATION & 47.13 \\ 
			RELIGION &47.13 \\
			POLITICS & 47.22\\
			SLOGANS & 47.13\\
			
			Combined Lists & 47.03\\
			\hline
			PERSON & 46.09 \\ 
			Various Entities&45.71\\
			\hline
		\end{tabular}
	\end{center}
	\caption{Baseline Model performance on Development set}
	\label{table:modelA}
	\end{table}

			



\subsection{Bert Model Results}
When fine-tuning the BERT model, we tried various approaches with the dataset. We tried using the raw dataset as well as a pre-processed one. Although pre-processing (with the techniques that we mentioned in Section~\ref{sssec:prepro}) improved results over the raw dataset, when we applied the mapping and the named entity recognition techniques we observed that pre-processing did not help achieve better results. The results are presented in Table~\ref{table:testbert}.

\begin{table}[htbp]
\begin{center}
		\begin{tabular}{|c|c|c|} 
			\hline
			Technique &Development Set $F_{1}$\\
			\hline
			Baseline Model & 46.37 \\ 
			BERT raw &51.14 \\
			BERT Pre-processed & 56.44 \\
			BERT Various Entities & 54.09 \\
			BERT Entity Person & 56.91 \\
			BERT Entity Person Pre-processed & 52.39 \\
			BERT Lists (all lists combined) & 57.85 \\
			BERT Lists Pre-processed& 55.03 \\
			\hline
		\end{tabular}
	\caption{BERT effectiveness on different instances of the dataset}
	\label{table:testbert}
	\end{center}
	\end{table}

\subsection{Final Submission Results}
By examining the results of our BERT models, we concluded that the best results came with mapping the dataset with the NATION, RELIGION, and POLITICS labels. The second best approach was with the PERSON tag that outperformed our best model in the Bandwagon, Flag Waving, Labeling, and Cliches categories. Our official submission to the competition ranked our team to the 10th place from 32 teams. The results of our model are shown in Table~\ref{table:bertfinal}.

\begin{table}[htpb]
\begin{center}
		\begin{tabular}{ |c|c|c|c|} 
			\hline
			\bf Label & \bf Test Set $F_{1}$  \\
			\hline
			
			Loaded\_Language & 73.70 \\ 
			Name\_Calling,Labeling	 & 71.40 \\ 
			Repetition & 20.10 \\ 
			Doubt & 59.15 \\ 
			Exaggeration,Minimisation	 & 28.23 \\
			Appeal\_to\_fear-prejudice	& 33.33 \\
			Flag-Waving	& 58.94 \\
			Causal\_Oversimplification	&26.23 \\
			Appeal\_to\_Authority	&44.44 \\
			Slogans	&34.78 \\
			Black-and-White\_Fallacy	&33.33\\
			Whataboutism,Straw\_Men	&17.77 \\
			Thought-terminating\_Cliches	&27.02 \\
			Bandwagon,Reductio\_ad\_hitlerum &9.30\\
			\hline
			Overall micro-averaged $F_{1}$ &57.20 \\
			\hline
		\end{tabular}
	\caption{\label{table:bertfinal}Final submission results}
	\end{center}
\end{table}

\section{Conclusions}\label{sec:conclusions}
We presented a supervised learning model for classifying text fragments from news articles in thirteen propaganda categories. We used standard classification techniques as well as modern NLP models such as BERT. We examined the task from a sociological point of view and we tried to experiment with the fact that different entities of the same type can have the same value for propaganda classification. The results were promising and further experiments could improve them. 
			
	
\bibliographystyle{coling}
\bibliography{semeval2020}

\end{document}